\documentclass[review]{elsarticle}
\graphicspath{ {./figures/} }
\usepackage{hyperref}
\usepackage{float}
\usepackage{verbatim} 
\usepackage{apalike}
\restylefloat{figure}
\restylefloat{table}
\usepackage{amsmath}
\usepackage{mathtools}
\usepackage{amsfonts}
\usepackage{enumitem}
\usepackage{forest}
\usepackage{graphicx}
\usepackage[caption=false]{subfig}
\usepackage{multirow}

\journal{Intelligent Systems with Applications}

\biboptions{authoryear}

\begin{document}
\begin{frontmatter}

\begin{titlepage}
\begin{center}
\vspace*{1cm}

\textbf{ \large Process Mining Embeddings: Learning Vector Representations for Petri Nets}

\vspace{1.5cm}

Juan G. Colonna$^{a}$ (juancolonna@icomp.ufam.edu.br), Ahmed A. Fares$^{b,c}$ (ahmed.a.fares@inesctec.pt), Márcio Duarte$^c$ (marcio.s.duarte@inesctec.pt), Ricardo Sousa$^c$ (ricardo.t.sousa@inesctec.pt) \\

\hspace{10pt}

\begin{flushleft}
\small  
$^a$ Computing Institute (IComp), Federal University of Amazonas (UFAM), Brazil \\
$^b$ Faculty of Engineering - University of Porto, Portugal \\
$^c$ INESC Tec, Porto, Portugal

\vspace{1cm}
\textbf{Corresponding Author:} \\
Juan G. Colonna \\
Computing Institute (IComp), Federal University of Amazonas (UFAM), Brazil \\
Tel: (+55) 92 98416-0589 \\
Email: juancolonna@icomp.ufam.edu.br

\end{flushleft}        
\end{center}
\end{titlepage}

\newpage
\textbf{ \large Highlights}

\begin{enumerate}
    \item \textbf{A Novel Approach for Process Model Embeddings:} This research introduces PetriNet2Vec, a new method inspired by Natural Language Processing (NLP) to create embedding vectors for process models represented as Petri nets.

    \item \textbf{Learning from Sentences, Understanding Processes:} PetriNet2Vec treats process paths like sentences in documents, allowing it to leverage NLP techniques for process mining tasks.

    \item \textbf{Unsupervised Learning for Robust Embeddings:} PetriNet2Vec learns effectively through unsupervised training with negative sampling, capturing the intricacies of various process models.

    \item \textbf{Empowering Downstream Tasks:} The learned embeddings prove valuable for process classification and retrieval, which are essential tasks in process mining.

    \item \textbf{Real-World Applications:} This study demonstrates success with a simulated dataset and proposes future work involving real-world data and incorporating broader task context for improved process discovery.
\end{enumerate}

\newpage
\title{Process Mining Embeddings: Learning Vector Representations for Petri Nets}

\author[label1]{Juan G. Colonna \corref{cor1}}
\ead{juancolonna@icomp.ufam.edu.br}

\author[label2,label3]{Ahmed A. Fares}
\ead{ahmed.a.fares@inesctec.pt}

\author[label3]{Márcio Duarte}
\ead{marcio.s.duarte@inesctec.pt}

\author[label3]{Ricardo Sousa}
\ead{ricardo.t.sousa@inesctec.pt}

\cortext[cor1]{Corresponding author.}
\address[label1]{Computing Institute (IComp), Federal University of Amazonas (UFAM), Brazil}
\address[label2]{Faculty of Engineering - University of Porto, Portugal}
\address[label3]{INESC Tec, Porto, Portugal}

\begin{abstract}
Process Mining offers a powerful framework for uncovering, analyzing, and optimizing real-world business processes. Petri nets provide a versatile means of modeling process behavior. However, traditional methods often struggle to effectively compare complex Petri nets, hindering their potential for process enhancement. To address this challenge, we introduce PetriNet2Vec, an unsupervised methodology inspired by Doc2Vec. This approach converts Petri nets into embedding vectors, facilitating the comparison, clustering, and classification of process models. We validated our approach using the PDC Dataset, comprising 96 diverse Petri net models. The results demonstrate that PetriNet2Vec effectively captures the structural properties of process models, enabling accurate process classification and efficient process retrieval. Specifically, our findings highlight the utility of the learned embeddings in two key downstream tasks: process classification and process retrieval. In process classification, the embeddings allowed for accurate categorization of process models based on their structural properties. In process retrieval, the embeddings enabled efficient retrieval of similar process models using cosine distance. These results demonstrate the potential of PetriNet2Vec to significantly enhance process mining capabilities.
\end{abstract}


\end{frontmatter}

\section{Introduction}

Modern business process models exhibit a level of complexity that renders traditional analysis tools insufficient for comprehensive understanding and optimization~\citep{recker_business_2009}. This challenge demands innovative approaches capable of extracting complex relationships governing process behavior.

Process discovery techniques leverage event data to extract, analyze, and visualize the actual execution of business processes. Generally, a data-driven approach with advanced algorithms is used to construct models, such as Petri nets, that accurately capture sequences of activities, \textit{i.e.} transitions (activity), and places (state)~\citep{Murata1989}. Petri nets are mathematical and visual modeling tools used to describe distributed systems. These networks allow us to assess process performance, identify bottlenecks, and uncover hidden patterns through conformance-checking techniques. Ultimately, facilitating process enhancement and optimization.

Traditional process mining core capabilities often fall short when faced with the scale and complexity of modern processes~\citep{ganesh_information_2023}. They may struggle to handle the vast amounts of data generated and lack the sophistication to uncover complex patterns that machine learning techniques could reveal~\citep{ana_rocio_cardenas_maita_systematic_2018}.

Embedding vectors are numerical representations of objects or concepts in a continuous vector space~\citep{jurafsky_speech_nodate}. Commonly used in Natural Language Processing (NLP) related tasks, which often involve machine learning algorithms. Embedding vectors capture semantic relationships between entities. In process mining, embeddings can provide a powerful means of representing complex process structures and relationships within process models~\citep{Pismerov2023}. For example, we can embed individual activities, control flow structures, or entire process models in vector representations. This enables sophisticated similarity analysis, identification of analogous patterns, and application of predictive modeling techniques that would be difficult to achieve with traditional process representations.

\subsection{Problem statement}

Given a set $\mathbb{M} = \{M_1, M_2, \dots, M_n \}$ of $n$ Process Models, where each process is represented by a Petri net in PNML format, our goal is to learn a $d$-dimensional embedding vector $\mathbf{x} \in \mathbb{R}^d$ for each model $M_j \in \mathbb{M}$. The matrix representation of all models will be denoted as $X \in \mathbb{R}^{n \times d}$, with each row representing an embedding vector. Our objective is to capture the structural dependencies of sequential tasks within the processes, facilitating similarity comparisons between models using cosine distance on vector embedding space.

Furthermore, since each model in $\mathbb{M}$ comprises several sequences of tasks and the total number of unique tasks across all processes is denoted as $\mathbb{T} = \{t_1, t_2, \dots, t_k \}$, we also aim to learn an embedding vector $\mathbf{t}_i \in \mathbb{R}^d$ for each task. The matrix of all task embeddings will be denoted as $T \in \mathbb{R}^{k \times d}$, with each row representing an embedding vector. These task embedding vectors serve to capture the inherent characteristics of each task within the process sequences.

\subsection{Contributions and overview}

This work introduces PetriNet2Vec, a novel method inspired by NLP to create embedding vectors for process models represented as Petri nets. By treating process paths like sentences in documents, PetriNet2Vec leverages NLP techniques to capture the intricacies of various process models through unsupervised training with negative sampling. Our approach encodes both the structural information of process models in Petri net format and the individual tasks into compact vector representations (\textit{i.e.}, embedding vectors), facilitating downstream tasks such as similarity analysis, process retrieval, and classification of process properties. Demonstrated on a simulated dataset, this approach shows promise for real-world applications, with future work aimed at incorporating broader task contexts to enhance process discovery. Additionally, we have made the PetriNet2Vec Python package publicly available for the community, which can be installed via the Python Package Index (PyPI)\footnote{\texttt{pip install PetriNet2Vec}}.

The article is organized as follows: Section~\ref{sec:related_works} provides a comprehensive review of related works in process mining, highlighting the existing methods, unveiling their limitations, and comparing them with our proposed PetriNet2Vec method through applying embedding techniques. In Section~\ref{sec:background}, we present the theoretical background of learning vector embeddings (\textit{e.g.} \textit{word2vec}), followed by the clustering algorithm and dataset description. Section~\ref{sec:methodology} details our methodology for learning Petri net embeddings, including the transformation of Petri nets into intermediate representations and the training process of PetriNet2Vec. Section~\ref{sec:results} presents the results of our experiments, including cluster analysis, visual assessments, and examples of downstream tasks such as process classification and retrieval. Finally, Section~\ref{sec:conclusions} concludes the manuscript with a summary of our findings, limitations, and suggestions for future research directions.

~

\section{Related works}\label{sec:related_works}


Process comparison is crucial in conformance analysis, process enhancement, knowledge transfer, and process retrieval. Existing techniques for process comparison broadly fall into three categories: behavioral analysis, structural analysis, and task comparison~\citep{syukriilah_structural_2015}. Behavioral strategies focus on the order of activities within execution logs, while structural approaches analyze the process model as a graph. Task comparison offers a granular view, looking at relationships between individual activities~\citep{dijkman_similarity_2011}.


\subsection{Behavioral analysis}
Behavioral process comparison focuses on the execution sequences observed in event logs.~\cite{hutchison_process_2006} emphasize the value of direct log analysis in uncovering process behavior. These methods, while effective in extracting sequences of activities, face challenges with terminology variations. PetriNet2Vec embeds entire process models, effectively capturing structural dependencies and semantic similarities, thus reducing the impact of terminology variations and improving scalability in handling complex processes.
~\cite{dijkman_similarity_2011} introduce the concept of ``causal footprints'', which capture the temporal dependencies between activities to represent process behavior.~\cite{rinderle-ma_behavioral_2011} highlight relationships between exclusivity and strict order as key behavioral aspects.~\cite{van_dongen_measuring_2013} develop behavioral metrics for quantifying process similarities and differences.~\cite{sanchez-charles_process_2016} extend this by proposing a behavioral distance measure to compare process behavior holistically. 

Although useful, those methods can be limited by the complexity introduced by concurrency and loops~\citep{bubenko_short_2013}. PetriNet2Vec’s embedding approach captures both dependencies and concurrency more efficiently, providing a comprehensive representation that scales better with complex processes.

\subsection{Structural analysis}
Alternatively, structural methods examine the overall process model as a graph, analyzing elements like node relationships, paths, connectivity, and control flow. Standard techniques utilize edit graph distance to calculate the similarity between process graphs~\citep{dijkman_graph_2009}. More nuanced methods consider richer information, such as node types (e.g., AND/XOR) and sequential information~\citep{montani_knowledge-intensive_2015}. While these methods are effective approaches reliant on graph edit distance, they face computational scalability challenges with large graphs due to their NP-Hard complexity~\citep{zeng_comparing_2009}. By representing process models as embedding vectors, PetriNet2Vec overcomes scalability issues and encodes structural information more efficiently, allowing for faster and more accurate comparisons.

\cite{zhou_comprehensive_2019} enhance structural techniques by incorporating insights from execution logs and weighting graph edges based on their frequency. While valuable for analyzing overall process flow, such techniques may not fully capture fine-grained behavioral variations that arise from the diverse execution of process activities. PetriNet2Vec integrates both structural and behavioral details in its embeddings, enabling finer-grained analysis and capturing nuances that purely structural methods might miss.

\subsection{Task comparison}
Task comparison provides a detailed analysis of activity relationships within process models. Node-by-node techniques excel at identifying subtle differences, while block-based methods reveal more significant structural changes~\citep{li_measuring_2008,dijkman_similarity_2011,bae_process_2006,yan_fast_2012}. On the other side, PetriNet2Vec provides a unified embedding representation, facilitating both detailed node-level and holistic structural comparisons, offering a more balanced approach to process model comparison. 

Recent clustering approaches hold the potential for comparing localized process sections~\citep{peeva_grouping_2023}. However, these localized clustering methods may struggle when analyzing processes with highly variable flows or behavioral patterns but can struggle with highly complex and diverse process models. By learning embeddings for tasks and entire processes, PetriNet2Vec enhances clustering capabilities, effectively managing variability and complexity in process models.

In the domain of Similarity-based retrieval of semantic graphs,~\cite{Hoffmann2022} proposed an approach employing Graph Neural Networks (GNNs) for process mining, framing it as a graph matching challenge. Despite the inherent complexity of GNNs, necessitating larger datasets for effective training, their methodology shows significant promise, particularly in retrieval scenarios essential to Process-Oriented Case-Based Reasoning (POCBR) within smart manufacturing - a focus distinct from our current study. Nonetheless, in Section\ref{sec:downstream_tasks}, we comprehensively evaluate our methodology, specifically in the realm of semantic retrieval tasks for process models, utilizing \textit{PetriNet2Vec.} This assessment underscores the resilience and adaptability of our retrieval approach, highlighting its potential for broader applications in the field.

\subsection{Conclusions on Related Work}

While current techniques offer valuable insights, their limitations highlight the need for a method that provides a more holistic view of process execution, considering both behavioral patterns and the broader process structure. Our proposed approach effectively addresses these needs. It considers the entire process and its structure, mirroring structural approaches, and enables the comparison of multiple processes and clusters, capturing the strengths of behavioral approaches.

PetriNet2Vec's comprehensive embedding approach offers a unified representation, facilitating both detailed node-level and task sequence modeling. This dual capability enhances the analysis and retrieval of complex process models, overcoming limitations of existing methods and demonstrating higher accuracy and scalability in clustering and classification tasks.

\section{Background}\label{sec:background}

\subsection{Learning embeddings with \textit{doc2vec} and \textit{graph2vec}}

\begin{figure}[htp]
\centering
\subfloat[\textit{Word2vec}.\label{fig:word2vec}]{%
\includegraphics[width=0.33\textwidth]{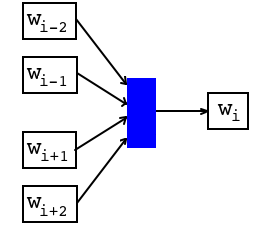}%
}\hfil
\subfloat[\textit{Doc2vec}.\label{fig:doc2vec}]{%
\includegraphics[width=0.33\textwidth]{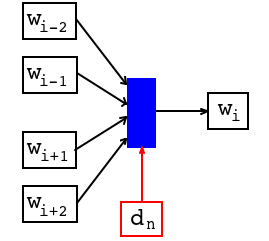}%
}\hfil
\subfloat[\textit{Graph2vec}.\label{fig:graph2vec}]{%
\includegraphics[width=0.33\textwidth]{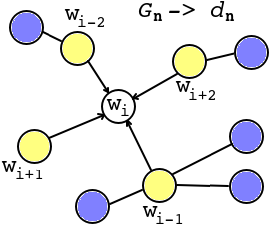}%
}
\caption{Subfigure~\ref{fig:word2vec} depicts the CBOW approach for \textit{word2vec}. Subfigure~\ref{fig:doc2vec} demonstrates the incorporation of the document ID from which the words were sampled. Subfigure~\ref{fig:graph2vec} illustrates a graph where the yellow nodes represent the `context' of node $w_i$, while adhering to the same nomenclature used in \textit{doc2vec} for the words within the context of $w_i$.}
\label{fig:background}
\end{figure}

The \textit{doc2vec} methodology~\citep{Mikolov2014}, an extension of the \textit{word2vec} concept, has become popular for enabling the simultaneous learning of embedding vectors for both documents and individual words. In \textit{word2vec}~\citep{Mikolov2013}, a word is predicted based on the words in its context. Figure~\ref{fig:background} illustrates the functioning of the CBOW (Continuous Bag-of-Words) approach. While this figure is useful for illustrating the \textit{word2vec} concept, we can reinterpret the optimization objective function during training as a binary classifier, predicting $P(1|w_i,w_c)$ when the word $w_i$ is part of the word context $w_c = \{ \dots, w_{i-2}, w_{i-1}, w_{i+1}, w_{i+2}, \dots\}$. Thus, if $w_i$ is not in the context $w_c$, then we have $P(0|w_i,w_c)$. Then, during training, the \textit{negative sampling} strategy generates positive and negative example pairs $(w_i,w_c)$.

Extending this formulation to learn embeddings for documents is straightforward. We can add a vector $d_n$ to the context $w_c$, treating it as a new word representing the entire document (Figure~\ref{fig:doc2vec}). Thus, the optimization objective that allows learning an embedding for this `new word' becomes predicting $P(1|w_i,w_c,d_n)$ when the word $w_i$ belongs to the context $w_c$ and has also been sampled from the document $d_n$, otherwise $P(0|w_i,w_c,d_n)$ when $w_i$ is not present in the document $d_n$.

The \textit{graph2vec} algorithm is a powerful formulation that allows us to learn to embed vectors for graph representation~\citep{Annamalai2017}. Given a graph $G_n=(V,N)$ consisting of a set of vertices $V$ and edges $N$, \textit{graph2vec} naturally extends the principles of \textit{doc2vec}. By analogy, each node in the graph can be seen as a word $V_i \equiv w_i$, with the neighboring nodes $V_c \equiv w_c$ directly connected to $V_i$, serving as its context. Consequently, if each node is associated with a token $t_k$, as depicted in Figure~\ref{fig:graph2vec}, we can concurrently learn embeddings for the entire graph $h_n$ and for each individual node.

This approach outperforms other graph embedding techniques because it
trains on various graph samples and incorporates neighborhood information for every node. Thus, it helps the neural network learn complex graph structures more effectively. As a result, \textit{graph2vec} embeddings can accurately capture similarities between graphs with similar structures, making them highly useful across different downstream applications~\citep{Cai2018}.

\subsection{Cluster algorithm}

As discussed in the previous sections, one of the objectives of this work is to learn embedding for each Petri net model and group them by similarity. For this purpose, we adopted a version of the Hierarchical Density-Based Spatial Clustering of Applications with Noise (HDBSCAN) clustering algorithm with cosine distance~\citep{mcinnes2017hdbscan}.

HDBSCAN stands out as a sophisticated clustering algorithm capable of effectively identifying clusters of various shapes and densities. Its hierarchical clustering methodology allows it to autonomously discern clusters at various density levels without prior knowledge of the ideal cluster count, providing flexibility in estimating the number of clusters present. Furthermore, thanks to implementing cosine distance, HDBSCAN demonstrates proficiency in handling high-dimensional data spaces, such as those found in text data or feature vectors.

Cosine similarity is a robust measure for clustering sparse data commonly encountered in vector embedding, accommodating fluctuations in vector magnitudes. This characteristic makes it particularly suitable for tasks where traditional techniques like k-means fail, especially when dealing with non-Euclidean distances such as cosine similarity. Unlike k-means, which faces challenges in combining the average centroid calculation with cosine distance due to their inherent incompatibility, HDBSCAN with cosine distance presents a versatile and effective solution for clustering sparse, high-dimensional data, positioning it as the chosen method for our application.

The output of HDBSCAN can be combined with the Silhouette score to evaluate the quality of formed clusters when ground truth labels are unavailable. This coefficient measures how well each data point fits into its assigned cluster, with values ranging from -1 to 1. A score near 1 indicates that data points are well within their clusters and far from others, while negative values close to -1 suggest poor cluster fit. Scores around 0 imply potential cluster overlap. Additionally, the average Silhouette score serves as a global measure, providing insights into the overall quality of the formed clusters.

\subsection{Process models dataset}\label{sec:dataset}

Our research leverages the PDC Dataset~\citep{PDC2023}, specifically curated for the Process Discovery Contest. This dataset encompasses a total of 96 Petri net models stored in PNML format. As a specialized class of models, Petri nets present a versatile range of configurations, rendering them indispensable for evaluating process discovery algorithms and techniques.

The dataset is derived from a base model named $\mathsf{pdc2023\_000000.pnml}$ and spans various configurations denoted by specific letters from A to F, such as $\mathsf{pdc2023\_ABCDEF.pnml}$. Each letter stands for a different configuration parameter:
\begin{enumerate}[label=\Alph*:]
    \item Dependent Tasks (Long-term Dependencies), configured as 0 (No) or 1 (Yes). If Yes, bypass connections are added to shortcut long-term dependent transitions;    
    \item Loops, configured as 0 (No), 1 (Simple), or 2 (Complex), determining the treatment of transitions initiating loops and shortcuts between the loop and main branches;
    \item OR constructs, configured as 0 (No) or 1 (Yes), controlling transitions involving only inputs or generating only outputs for OR-joins and OR-splits;
    \item Routing Constructs (Invisible Tasks), configured as 0 (No) or 1 (Yes), making certain transitions invisible when set to Yes;
    \item Optional Tasks, configured as 0 (No) or 1 (Yes), enabling the skipping of visible transitions with the addition of invisible transitions when set to Yes; and
    \item Duplicate Tasks (Recurrent Activities), configured as 0 (No) or 1 (Yes), involving the relabeling of transitions to account for duplicate tasks when set to Yes.
\end{enumerate}

All candidate models were systematically generated following a sequential rule that applies each parameter to the base model. In essence, the characters in this naming convention serve as flags that, when activated, apply these specific rules to the process model. For instance, the first four candidate models are named $\mathsf{pdc2023\_000000.pnml}$,  $\mathsf{pdc2023\_000001.pnml}$, $\mathsf{pdc2023\_000010.pnml}$, and $\mathsf{pdc2023\_000011.pnml}$. This nomenclature signifies that the second, third, and fourth models are variations of the base model, incorporating Duplicate Tasks (F), Optional Tasks (E), or both (F and E), respectively. Subsequently, the next configuration D ($\mathsf{pdc2023\_000100.pnml}$) is activated, and the procedure repeats. 

\begin{figure}[t]
\centering
\includegraphics[width=\textwidth]{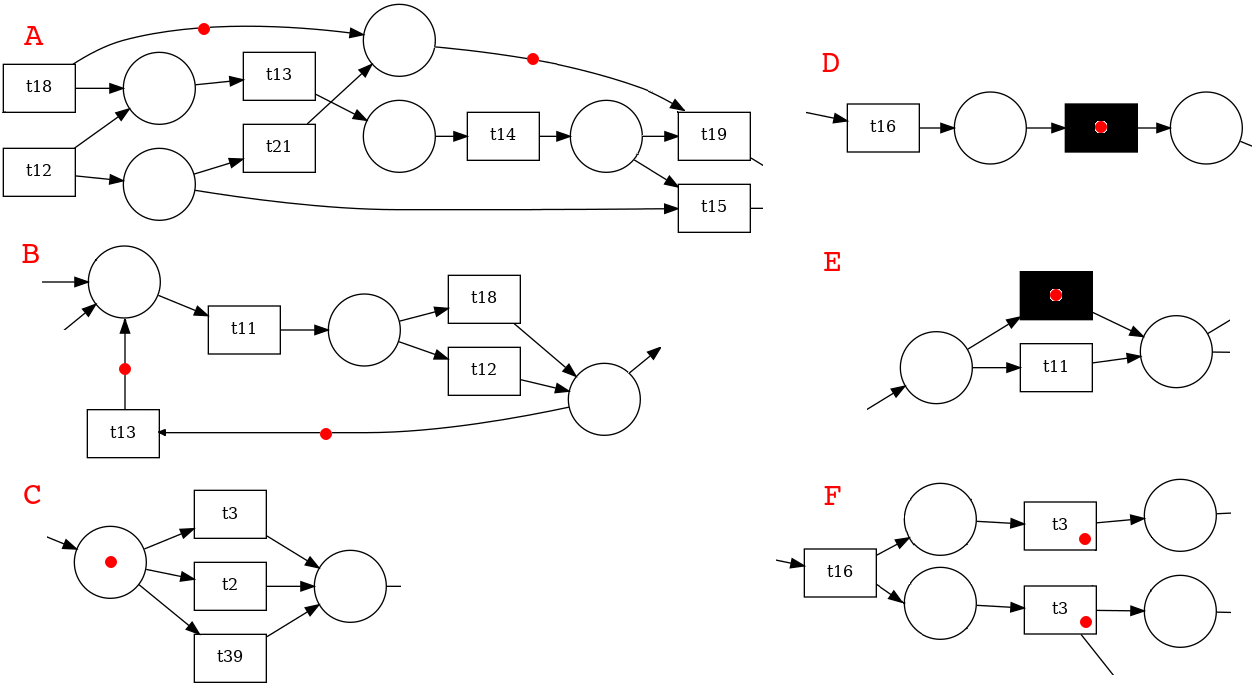}
\caption{Illustration of the rules applied to a process model. Red dots indicate the effect caused by each rule. (A) shows a bypass connection, (B) shows a loop, (C) shows an OR-construct, (D) shows an invisible task, (E) shows an optional task, and (F) shows a duplicated task and demonstrates an AND split.} \label{fig:rules}
\end{figure}

Figure~\ref{fig:rules} illustrates through Petri net segments where these rules have been applied. This sequential generation process encompasses all possible combinations, where the $2^5$ binary combinations are multiplied by the three configurations of rule B, resulting in a total of $2^5 * 3 = 96$ model variations. Understanding this generation procedure is fundamental to grasping how our learning algorithm identifies clusters with similar models generated by comparable combinations of these rules.

\section{Methodology for Learning Petri Net Embeddings}\label{sec:methodology}

\begin{figure}[htpb]
\centering
\includegraphics[width=1.\textwidth]{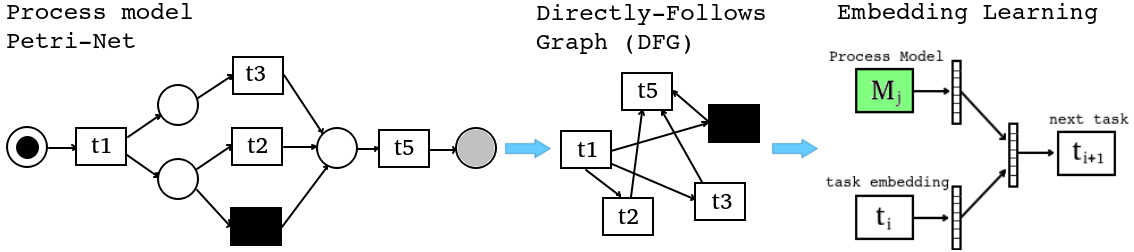}
\caption{Proposed methodology. On the left, a Petri net representation of a process model; in the middle, an equivalent representation as a directed graph of transitions; on the right, a Distributed Memory algorithm used for jointly learning embeddings for the process and tasks.} \label{fig:methodology}
\end{figure}

Consider the Petri net illustrated on the left side of Figure~\ref{fig:methodology} as an example of a process model. In this representation, each box represents a transition, and the circles represent states. The black dot is a token that traverses this network from left to right, visiting all states along a path. A transition must be completed to progress from one state to the next. Therefore, transitions represent tasks that need to be accomplished. Transitions are pivotal elements in these representations as they denote the activities that constitute the entire process. Transitions can be labeled using natural language. Here, for consistency and simplicity, we have labeled them using IDs, namely as $t_1$, $t_2$, and so forth.

In a Petri net, paths are represented by directed arrows, which also denote temporal dependencies. For instance, the transition $t_3$ cannot be initiated until the transition $t_1$ is completed. However, some paths can be executed in parallel; for example, transition $t_3$ can be executed concurrently with $t_2$. This indicates that a predecessor transition, like $t_1$, acts like an AND split, where both resultant paths are taken. Conversely, if a state splits into two paths, it is interpreted as an XOR split, meaning that only one of the two paths is chosen, such as $t_2$ or the black transition, but never both. Finally, the black box is a \textit{Silent} transition and functioning as a wildcard. In this example, the black box enables skip connections in the model \textit{i.e.} permitting optionally execution of $t_2$.

Our hypothesis is that since every transition can be represented in natural language, and there is a mandatory order dependency between these transitions, then a path inside the model can be considered equivalent to a sentence in a text document. Therefore, the set of all possible paths inside the model is equivalent to a text paragraph. Hence, without loss of generality, we can assume that the \textit{doc2vec} methodology can be employed here to learn a vector embedding for each process model in our dataset. Indeed, we can also jointly learn a vector embedding for every task by considering the set of all tasks in all models.

The embedding vectors generated by \textit{doc2vec} carry semantic meanings about documents and words. Here, these vectors will convey information and meaning about the process model and about every task across all models. Thus, those vectors may be useful for downstream tasks, such as model comparison, clustering, or classification. Therefore, by analogy, we named our methodology as PetriNet2Vec.

Despite its simplicity, a key step in our methodology is mapping a process model into an equivalent representation suitable for algorithm training and embedding vector learning. Here, we introduce the concept of \textit{graph2vec}. Prior to training, we generate a document list containing tasks. To achieve this, we transform each Petri net into an intermediate representation resembling a Directly-Follows Graph (DFG). DFGs offer a popular and comprehensible format for representing business processes in process mining. Figure~\ref{fig:methodology} (center) illustrates a DFG derived from the Petri net on the left. In this graph, we remove places and directly connect tasks, forming a classic directed graph where each node represents a task. Using that DFG, we construct tuples $(t_i, t_{i+1}, m_j)$, where $m_j$ is a unique identifier for the $j$-th process model in the dataset. 

To train PetriNet2Vec, we represent a model as the set of all possible transition pairs within it. Furthermore, we adopted the Distributed Memory (DM) algorithm implemented by the Gensim~\citep{gensim} library in Python, as illustrated on the right side of Figure~\ref{fig:methodology}. The diversity of process models in the training dataset ensures that some pairs of transitions are more frequent in some process models than others, and some transition pairs may even exist in some models but not in others. Therefore, by training using the described tuples, we can ensure that the learned vectors represent every model uniquely, carrying information about their structure. During training, we aim to maximize the probability of task $t_{i+1}$ being in the context of task $t_i$ and model $m_j$. We train on the entire dataset, maximizing the average probability loss:
\begin{equation}\label{equ:maximize}
    \text{maximaze:} ~ \frac{1}{M} \sum_{j}^M \frac{1}{T} \sum_{i}^{T} \log P(t_{i+1}|t_{i},m_j),
\end{equation}
\noindent where $T$ represents the set of all tasks and $M$ denotes the set of all models involved.

This training method does not require any model labels or supervision; in other words, this embedding training is unsupervised. To improve the separability between the learned vectors, we adopted the negative sampling strategy~\citep{Mikolov2013}. With this approach, in every training epoch, some pairs are randomly sampled from other models to form negative tuples for training that do not exist in the current model being trained. With negative sampling, the maximization objective becomes $P(1|t_{i+1},t_{i},m_j)$, \textit{i.e.}, predict 1 when the task $t_{i+1}$ is in the context of task $t_i$ and model $m_j$. Nevertheless, all these sampling and training epochs occur in parallel for speed and efficiency improvements.

\begin{figure}[t]
\centering
\subfloat[Methodoly 1 (`None').\label{fig:histogram1}]{%
\includegraphics[width=0.98\textwidth]{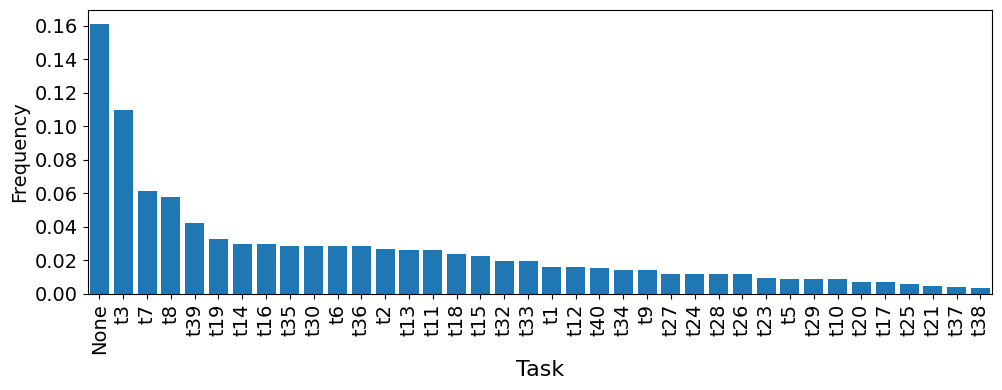}%
}\hfil
\subfloat[Methodoly 2.\label{fig:histogram2}]{%
\includegraphics[width=0.98\textwidth]{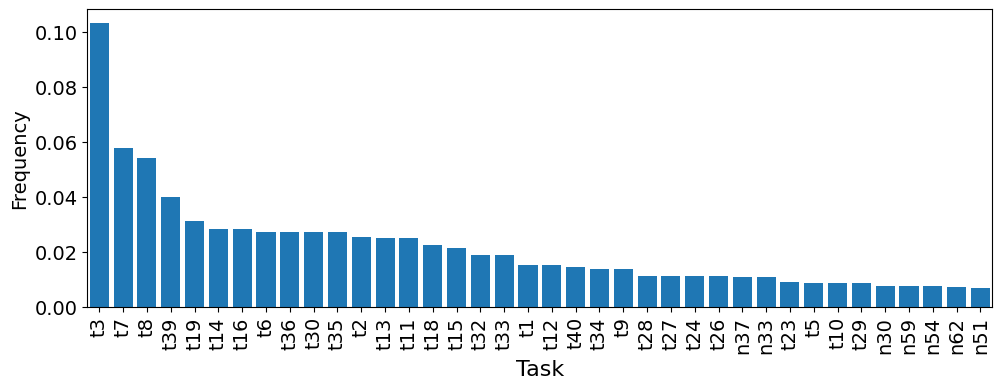}%
}
\caption{Caption for this figure with two images}
\label{fig:histograms}
\end{figure}

\section{Results}\label{sec:results}

\subsection{Cluster analysis}

As previously discussed, specific tasks within our Petri net models may be unspecified, denoted by black boxes. We face two choices in developing our methodology: we can adopt a generic and uniform name for all black boxes in all models, for instance, using the `None' token, akin to an out-of-vocabulary token in \textit{word2vec}, or uniquely label each black box according to its position in the network, referred to as `n$_{x}$' transitions from Figure~\ref{fig:histograms}. These alternatives for constructing the task dictionary (also known as the dictionary of tokens) influence the embedding vectors we aim to learn during PetriNet2Vec model training. Thus, our initial inquiry delves into how the quality of the clusters is impacted under each scenario.

Figure~\ref{fig:histograms} illustrates the histograms depicting task distributions across all models when employing the `None' token versus individually naming black boxes. A comparative analysis reveals stark differences in task frequencies. Utilizing unique names for black boxes necessitates the Distributed Memory (DM) algorithm to learn a more extensive set of embedding vectors, given the expanded token dictionary. While this might offer an advantage in distinguishing structures between models, our relatively modest dataset for training implies that the resultant vectors may suffer in quality. In essence, attempting to learn more embedding vectors without sufficient sample Petri net models compromises the final cluster quality across models. Therefore, due to these constraints, methodology 1, which entails a more straightforward approach with a smaller token dictionary, has a better chance of converging to more discriminative embedding vectors. This assertion finds validation in the results presented in Table~\ref{tab:hdbscan1}.

Tables~\ref{tab:hdbscan1} and~\ref{tab:hdbscan2} compare the silhouette scores for the different methodologies. The leftmost column indicates the size of embeddings, $d_n$, while the adjacent columns vary the only hyperparameter, `minimum cluster size', within the HDBSCAN clustering algorithm using cosine distance. Lower minimum cluster sizes increase the likelihood of forming smaller clusters, whereas higher values tend to produce larger, amalgamated clusters. Since the silhouette score is sensitive to cluster size, it is crucial to adjust this parameter within reasonable bounds. 
We determined that an acceptable range is between 2 and 6, as these values balance the need for distinct, well-defined clusters with the necessity of maintaining meaningful groupings, making them effective for process retrieval and clustering.

The silhouette scores were computed by running each methodology 30 times, with 500 training epochs for each investigated embedding size. The central values in the table represent the averages across these runs, while the values in parentheses indicate the standard deviations. Consistently, Methodology 1 (Table~\ref{tab:hdbscan1}) yielded superior silhouette scores compared to Methodology 2 (Table~\ref{tab:hdbscan2}), indicating more cohesive clusters. Optimal results were attained with $d_n = [4,8]$ and a minimum cluster size of $[3,4,5,6]$. Although $d_8$ does not have the highest score, it demonstrates the best compromise by achieving minimal variance, indicating more consistent cluster formation across the 30 runs.

We applied the t-test with a significance level of $p=0.01$ for two independent samples to determine if there is a tie between the mean silhouette scores of the groups. This test evaluates whether the observed difference between the means is statistically significant, allowing us to detect ties between hyperparameter combinations. The null hypothesis states that the means of the two groups are equal, while the alternative hypothesis posits that the means are different. We tested the scores resulting from $d_8$ and a minimum cluster size of 4 against all other pairs of hyperparameter combinations. The bold values in Table~\ref{tab:hdbscan1} indicate where the null hypothesis was accepted, showing no significant difference in means. Consequently, we decided to explore Methodology 1, employing embeddings of size $d_8$ and a minimum cluster size of 4.

\begin{table}[t]
\footnotesize
\centering
\caption{Comparison of silhouette scores between Methodology 1 (`None'). Silhouette scores were averaged across 30 runs with 500 training epochs for each embedding size. Methodology 1 consistently yields superior results, particularly evident with embedding sizes of $d_n=[4, 8]$.}
\label{tab:hdbscan1}
\begin{tabular}{ccccccc}
\hline
      &  & \multicolumn{5}{c}{Minimum cluster size}                                                                \\ \cline{3-7} 
$d_n$ &  & 2           & 3                    & 4                    & 5                    & 6                    \\ \cline{1-1} \cline{3-7} 
4     &  & 0.64 (0.09) & \textbf{0.71 (0.07)} & \textbf{0.74 (0.07)} & \textbf{0.74 (0.07)} & \textbf{0.74 (0.06)} \\
8     &  & 0.7 (0.02)  & \textbf{0.71 (0.02)} & \textbf{0.72 (0.02)} & \textbf{0.73 (0.04)} & \textbf{0.73 (0.03)} \\
16    &  & 0.48 (0.01) & 0.48 (0.02)          & 0.37 (0.04)          & 0.36 (0.02)          & 0.37 (0.01)          \\
32    &  & 0.46 (0.01) & 0.4 (0.03)           & 0.33 (0.02)          & 0.35 (0.01)          & 0.35 (0.01)          \\ \hline
\end{tabular}
\end{table}

\begin{table}[t]
\footnotesize
\centering
\caption{Comparison of Silhouette Scores between Methodology 2. The silhouette scores were averaged across 30 runs with 500 training epochs for each embedding size. Methodology 2 showed worse results compared to Methodology 1.}
\label{tab:hdbscan2}
\begin{tabular}{ccccccc}
\hline
      &  & \multicolumn{5}{c}{Minimum cluster size}                            \\ \cline{3-7} 
$d_n$ &  & 2           & 3           & 4           & 5           & 6           \\ \cline{1-1} \cline{3-7} 
4     &  & 0.58 (0.07) & 0.62 (0.08) & 0.48 (0.08) & 0.51 (0.07) & 0.54 (0.05) \\
8     &  & 0.44 (0.04) & 0.45 (0.04) & 0.27 (0.05) & 0.26 (0.06) & 0.32 (0.04) \\
16    &  & 0.32 (0.02) & 0.31 (0.02) & 0.2 (0.02)  & 0.19 (0.02) & 0.17 (0.07) \\
32    &  & 0.3 (0.01)  & 0.29 (0.01) & 0.18 (0.01) & 0.17 (0.03) & 0.17 (0.01) \\ \hline
\end{tabular}
\end{table}

\subsection{Visual Cluster Assessment}

\begin{figure}[htpb]
\centering
\subfloat[Silhouette scores of formed clusters.\label{fig:shilloutte_models}]{%
\includegraphics[width=0.49\textwidth]{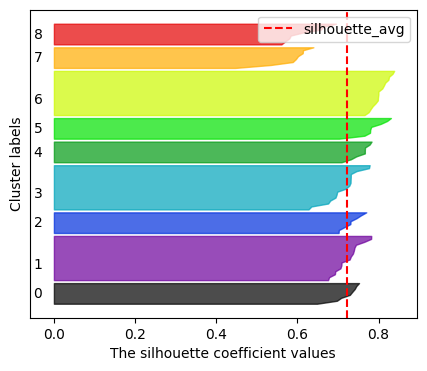}%
}\hfil
\subfloat[UMAP projection.\label{fig:UMAP_models}]{%
\includegraphics[width=0.49\textwidth]{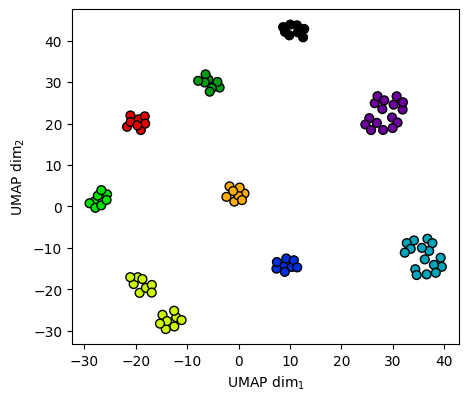}%
}
\caption{Visual assessment. Left: Silhouette plot of model clusters with average silhouette indicated. Right: UMAP projection of process models with cluster colors.}
\end{figure}

After determining the size of the embedding vectors and running the HDBSCAN algorithm with cosine distance, it was found that the models naturally cluster into nine groups. The silhouette plot depicted in Figure~\ref{fig:shilloutte_models} illustrates the distribution of models across these clusters. Higher silhouette values indicate better inner cluster cohesion. The dashed vertical red line represents the average silhouette across all clusters, measuring the overall quality of the formed groups. Clusters with more members appear wider. The smallest cluster size identified by HDBSCAN was eight, while the largest was seventeen.

Since the process models are represented by vectors of size 8, it is necessary to employ dimensionality reduction techniques to visualize them. We chose the UMAP technique to project these models into two dimensions, as shown in Figure~\ref{fig:UMAP_models}. In this scatter plot, each point represents a process model (a Petri net) from our dataset. The colors of the points correspond to the clusters identified in the Silhouette plot. We opted for UMAP for two reasons: its 2D projection is non-linear and can be performed using cosine distance. This figure demonstrates excellent quality and separation among the groups formed by the embedding vectors. However, it is important to note that we are observing a distorted representation of an 8-dimensional vector space, so some groups of points may not be well accommodated in 2D.

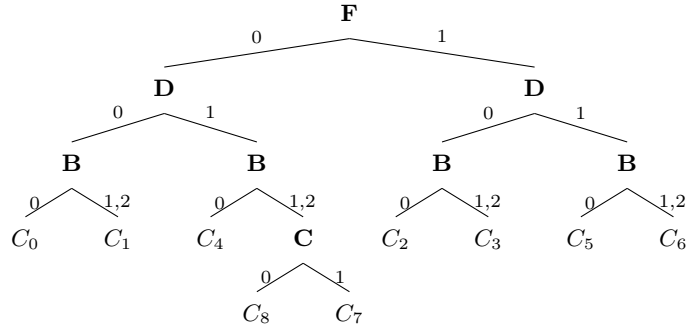
\begin{figure}[htpb]
\centering
\begin{forest}
  for tree={
    parent anchor=south,
    child anchor=north,
    if n children=0{
      font=\itshape
    }{},
    minimum width=1cm,
    minimum height=0.1cm,
    align=center,
  }
  [\textbf{F}
    [\textbf{D}, edge label={node[midway,left,above=0.5pt,font=\scriptsize]{0}}
      [\textbf{B}, edge label={node[midway,left,above=0.5pt,font=\scriptsize]{0}}
        [$C_0$, edge label={node[midway,left,font=\scriptsize]{0}}]
        [$C_1$, edge label={node[midway,right,font=\scriptsize]{1,2}}]
      ]
      [\textbf{B}, edge label={node[midway,right,above=0.5pt,font=\scriptsize]{1}}
        [$C_4$, edge label={node[midway,left,font=\scriptsize]{0}}]
        [\textbf{C}, edge label={node[midway,right,font=\scriptsize]{1,2}}
            [$C_8$, edge label={node[midway,left,font=\scriptsize]{0}}]
            [$C_7$, edge label={node[midway,right,font=\scriptsize]{1}}]
        ]
      ]
    ]
    [\textbf{D}, edge label={node[midway,right,above=1pt,font=\scriptsize]{1}}
      [\textbf{B}, edge label={node[midway,left,above,font=\scriptsize]{0}}
        [$C_2$, edge label={node[midway,left,font=\scriptsize]{0}}]
        [$C_3$, edge label={node[midway,right,font=\scriptsize]{1,2}}]
      ]
      [\textbf{B}, edge label={node[midway,right,above,font=\scriptsize]{1}}
        [$C_5$, edge label={node[midway,left,font=\scriptsize]{0}}]
        [$C_6$, edge label={node[midway,right,font=\scriptsize]{1,2}}]
      ]
    ]
  ]
\end{forest}
\caption{Decision Tree revealing common formation rules within clusters. The most discriminative rule \textbf{F}, which distinguishes between having or not having duplicated tasks, serves as the root of this tree.}\label{fig:tree}
\end{figure}

Upon closer examination within the cluster members, we observed emerging patterns related to the formation rules used to generate this dataset, as described in Section~\ref{sec:dataset}. For example, the process models within cluster $C_0$ include Petri nets\footnote{These are abbreviated names of the models; for instance, `pdc2023\_101010.pnml' is shortened to `101010'.}: `000000', `000010', `001000', `001010', `100000', `100010', `101000', `101010', all of which lack loops (rule $B=0$), invisible tasks (rule $D=0$), and duplicated tasks (rule $F=0$). We observed similar patterns in other clusters, characterized by different combinations of these rules. This raises an interesting question: Can we identify the formation rules common to all cluster members? To investigate this further, we fitted a decision tree using learning embedding  as feature vectors and the cluster identifier as labels. The resulting tree is shown in Figure~\ref{fig:tree}. This tree perfectly fits our clusters, revealing the common formation rules shared by all members of each formed cluster. For instance, all members of cluster $C_7$ lack duplicated tasks (rule $F=0$) but have invisible tasks (rule $D=1$), loops (rule $B=1$ or $B=2$), and OR-splits (rule $C=1$). This strong coherence among cluster members indicates that our methodology discovers the structural properties of the process models.

\begin{figure}[htpb]
\centering
\subfloat[Base model `000000' (no active rules).]{%
\includegraphics[width=\textwidth]{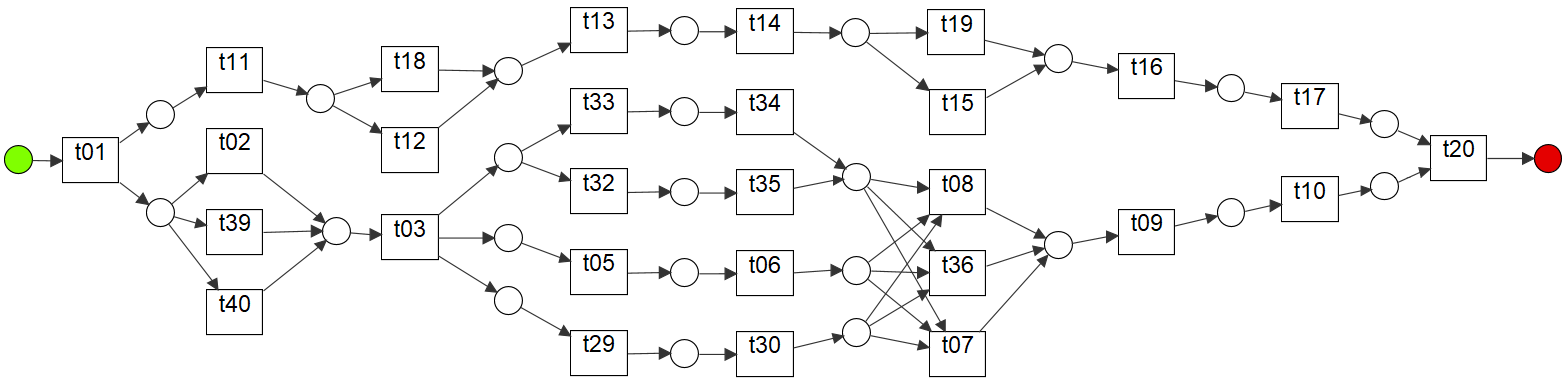}%
}\hfil
\subfloat[`000010' (optional tasks).]{%
\includegraphics[width=\textwidth]{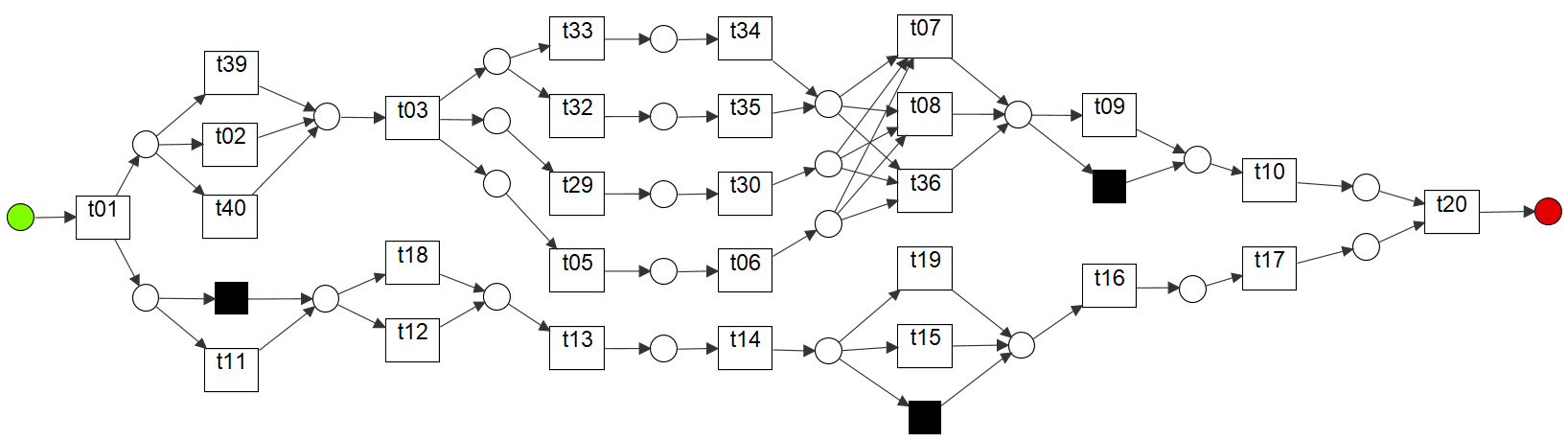}%
}\hfil
\subfloat[`001000' (OR-constructs).]{%
\includegraphics[width=\textwidth]{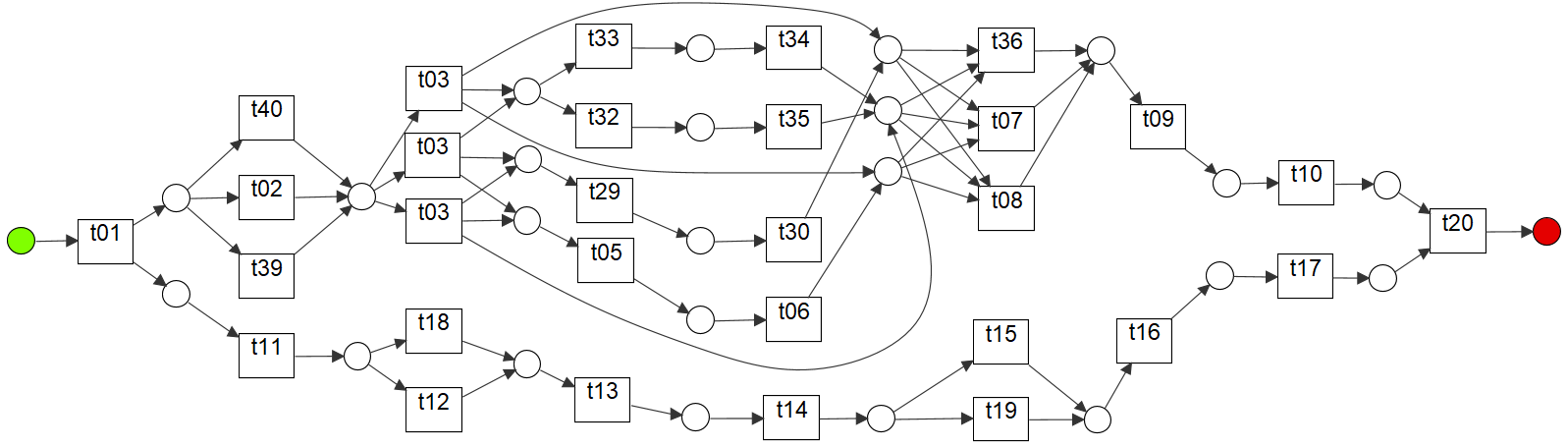}%
}\hfil
\subfloat[`001010' (optional tasks and OR-constructs).]{%
\includegraphics[width=\textwidth]{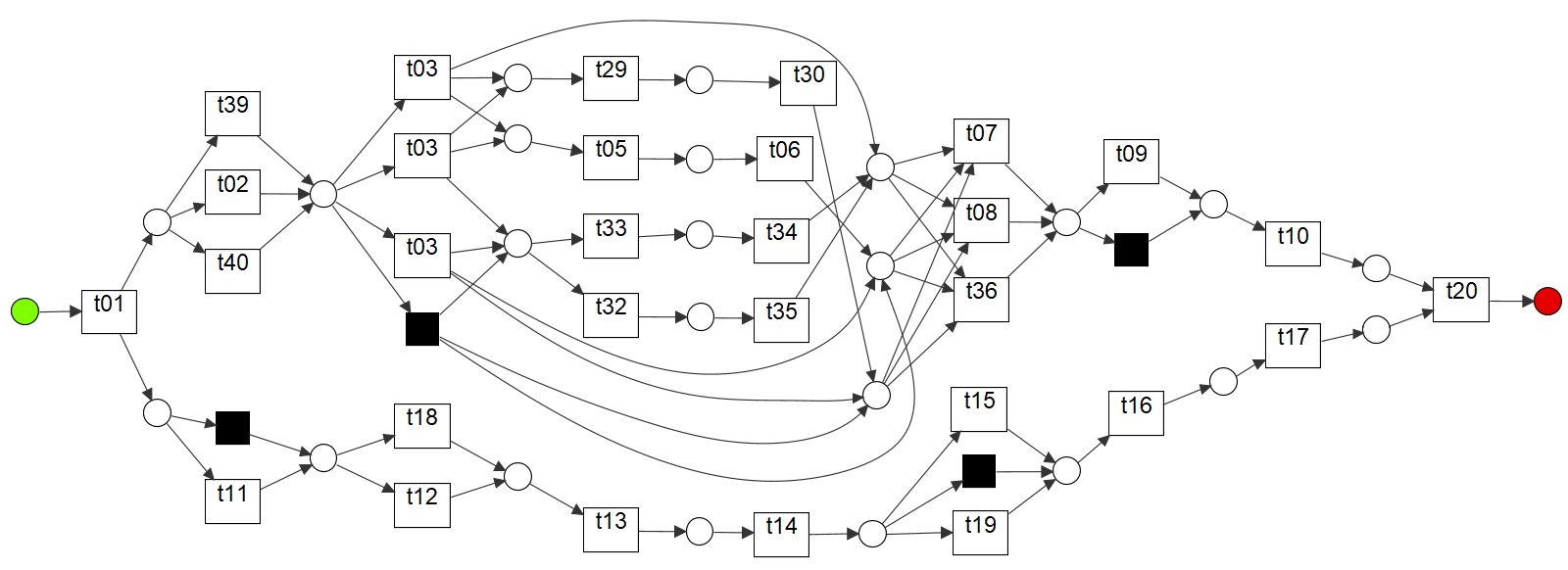}%
}
\caption{Process models members of cluster $C_0$.}
\label{fig:cluster_members}
\end{figure}

Finally, Figure~\ref{fig:cluster_members} displays the first four members of cluster $C_0$, including the base model and their neighbors. From these models, it can be observed that none of them contain loops, invisible tasks, or duplicated tasks.

\subsection{Expanding Our Cluster Analysis to Task Embeddings}

\begin{figure}[htpb]
\centering
\subfloat[Silhouette from task embeddings.\label{fig:silhouette_task}]{%
\includegraphics[width=0.49\textwidth]{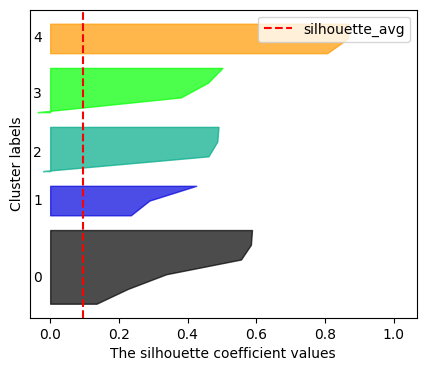}%
}\hfil
\subfloat[UMAP from task embeddings.\label{fig:UMAP_task}]{%
\includegraphics[width=0.49\textwidth]{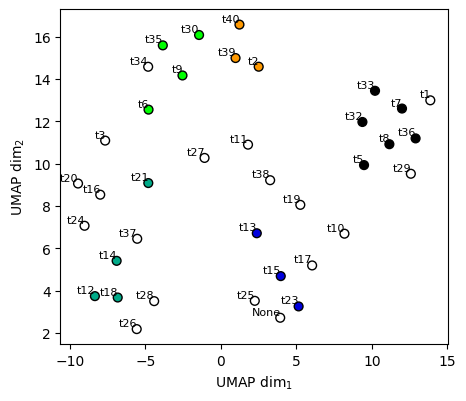}%
}
\caption{Visual assessment of task embeddings.}
\end{figure}

Until now, our analysis has primarily focused on the embeddings of models. However, during PetriNet2Vec training, we also learn task embeddings, meaning that each task depicted in Figure~\ref{fig:histogram1} is also mapped to an 8-dimensional embedding vector. Employing HDBSCAN on these task embeddings revealed the emergence of five clusters. Figure~\ref{fig:silhouette_task} displays the Silhouette plot, while Figure~\ref{fig:UMAP_task} illustrates the 2-dimensional projection of these task vectors. 

Unlike the clusters of models, the clusters of tasks exhibit lower quality, and some tasks were not associated with any cluster, indicated as white dots (or cluster noises). This occurred because the cosine similarity with their closest neighbor was considered too large by the HDBSCAN algorithm. Nonetheless, they still exhibit some degree of cohesion, and most importantly, they play a significant role during the training of model embeddings, granting PetriNet2Vec greater flexibility in accommodating those model embeddings.

Upon examining the scatter plot in Figure~\ref{fig:silhouette_task}, we observed a tendency for tasks sharing OR-splits, such as $t_2$, $t_{39}$, and $t_{40}$, to be grouped together, likely owing to their resemblance to text synonyms in \textit{doc2vec}. By analogy, tasks preceding an OR-split create alternate paths within the models, enabling the process to achieve the same subsequent state through diverse routes. 

Intriguingly, some white tasks are frequently subject to substitution by black boxes when the $D=1$ rule is applied. As these tasks are replaced, their occurrence diminishes in PNML files, consequently reducing their sampling during training. Moreover, infrequent tasks exist in models that are not necessarily substituted by black boxes. Analogously, these tasks mimic the behavior of less frequent words in \textit{doc2vec}. This becomes apparent when contrasting the histogram in Figure~\ref{fig:histogram1} with Figure~\ref{fig:UMAP_task}, where less frequent tasks are predominantly denoted as white dots. Nonetheless, while other analogies may be drawn, it is crucial to acknowledge that certain cluster artifacts may arise due to the inherent limitations of HDBSCAN.

\begin{figure}[htpb]
\centering
\includegraphics[width=.7\textwidth]{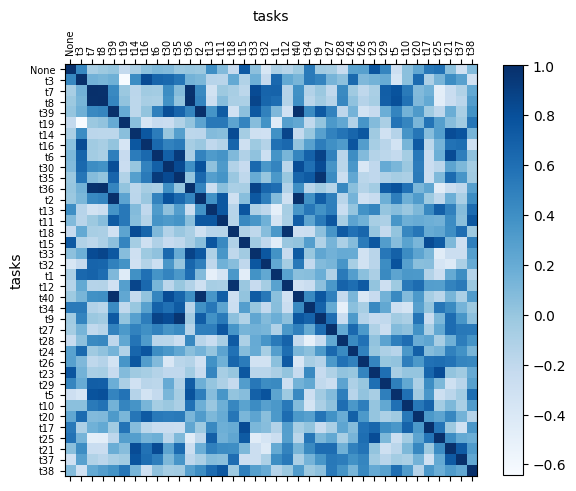}
\caption{Cosine similarity matrix illustrating relationships between task embeddings.} \label{fig:task_similarity_matrix}
\end{figure}

To extend our analysis beyond the clusters produced by HDBSCAN, we also examined the similarity matrix depicted in Figure~\ref{fig:task_similarity_matrix}. This matrix represents the cosine similarity between every pair of tasks. The stronger the color, the higher the similarity. From this matrix, we can address questions like: What are the most similar (or dissimilar) tasks to a given task in vector space? Since we use a task dictionary with only task IDs, giving them meaning is challenging. However, in a dataset with real business process models, these IDs are translated into interpretable tasks that must be accomplished to navigate the Petri net process flow. For instance, from this matrix, we can identify the set of tasks most similar to $t_{36}$ are $t_{7}$ and $t_8$, and the most dissimilar are $t_{13}$ and $t_{25}$. This highlights that $t_{13}$ and $t_{25}$ lie in a parallel branch to $t_{36}$.

\subsection{Examples of Downstream Tasks}\label{sec:downstream_tasks}

Once our methodology is complete and the PetriNet2Vec algorithm has been trained, we can tackle several downstream tasks using the learned embeddings, such as model or task query for similarity, model or task classification, and more. Also, we can explore what PetriNet2Vec algorithm has already learned by examining the relationships between task embeddings and model embeddings.

\begin{figure}[htpb]
\centering
\subfloat[Query model `010100'.]{%
\includegraphics[width=\textwidth]{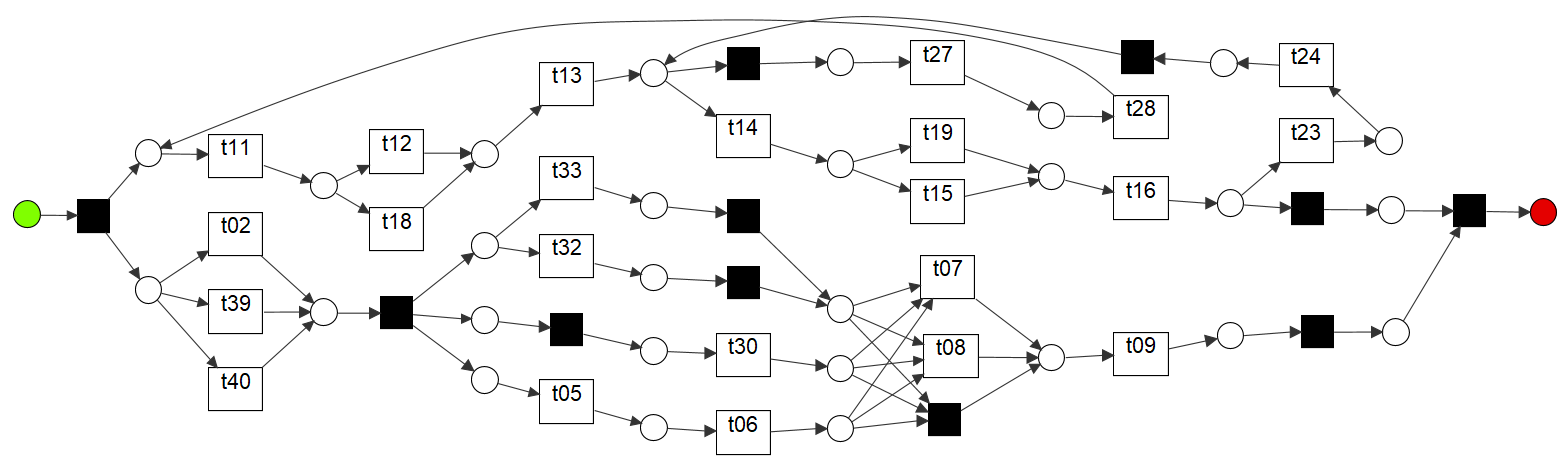}%
}\hfil
\subfloat[Answer model `010110'.]{%
\includegraphics[width=\textwidth]{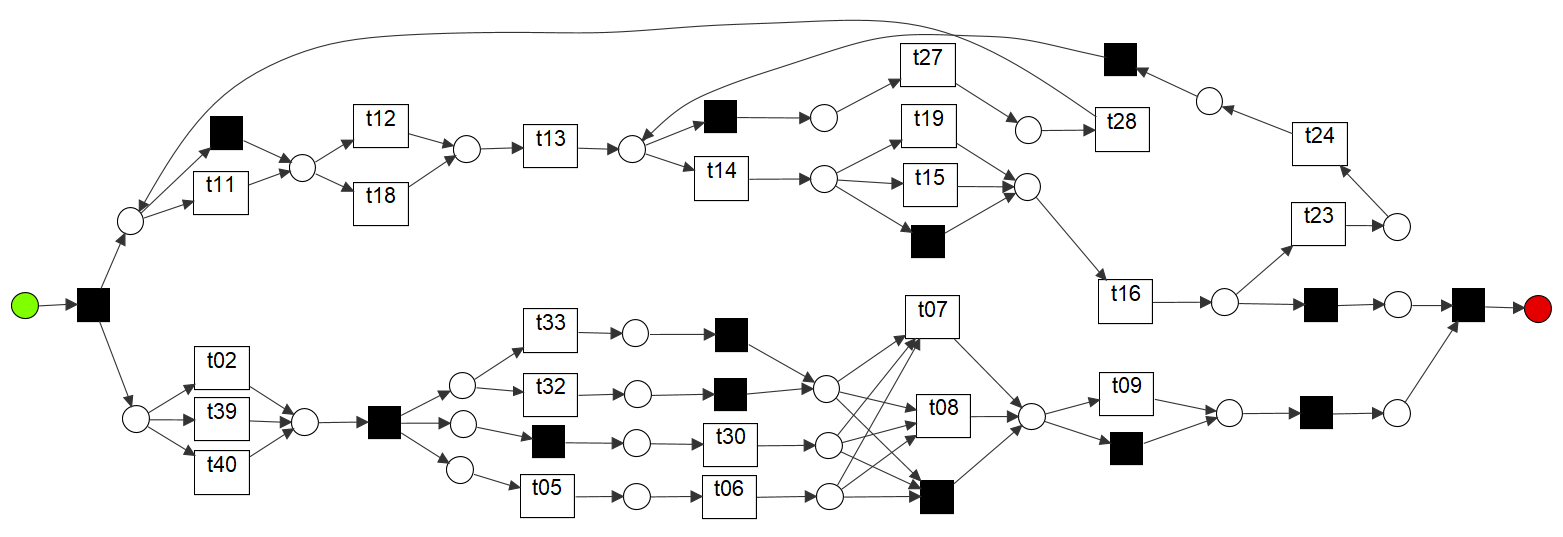}%
}
\caption{Example of solving a model query using cosine similarity. In this case, the red dots in (b) highlight the only two differences between the query and the returned models, providing a clear visual representation of the similarities and differences between them.}
\label{fig:model_query}
\end{figure}

Our first example of the downstream task is the  \textit{model query}, which involves randomly selecting a model, obtaining its embedding vector, and consulting the most similar model in the database by comparing the cosine distance between the query and each model represented by an embedding vector. Figure~\ref{fig:model_query} demonstrates this process retrieval task, where model 20 (`010100') was the query and model 22 (`010110') was the most similar recovered model. Upon comparing these two process models, we observe that the only difference is the activation of rule E, optional paths, in the returned model. This approach can also be adapted to return multiple answers and rank them accordingly.

\begin{figure}[htpb]
\centering
\includegraphics[width=1\textwidth]{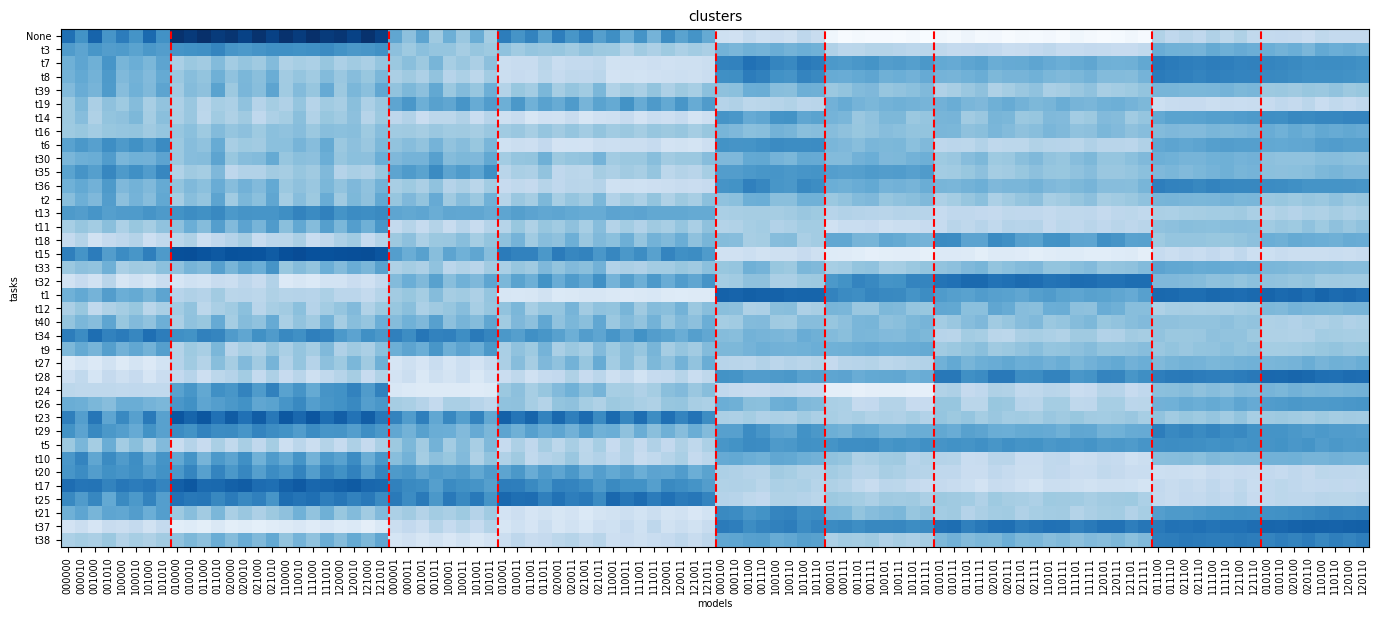}
\caption{Task and model embeddings' cosine similarity visualization. Models are grouped by clusters, separated by vertical dashed lines. This visualization highlights the relationships learned between tasks and models, facilitating the identification of similar processes and tasks within the business process model.} \label{fig:models_vs_tasks}
\end{figure}

Addressing the complexity and significance of querying business process models provides invaluable insights for large enterprises managing numerous processes. It enables them to pinpoint structurally similar processes, facilitating informed decisions such as merging or replacing process models. Moreover, this methodology can be extended to identifying similar tasks within processes. Figure~\ref{fig:models_vs_tasks} visually demonstrates an investigation of this nature, illustrating the similarity between each task embedding and each model embedding. In this figure, the models are ordered according to the clusters they formed during the application of HDBSCAN. Notably, certain task embeddings closely align with all model embeddings within the same cluster, unveiling compelling patterns. 

Nevertheless, interpreting task-model embeddings demands caution. Although they share the same embedding dimensions, they are different vector spaces. Consequently, the resulting score would not directly translate to a strict interpretation like model-to-model similarity. Nevertheless, it still offers valuable insights into the task's relevance to the model's knowledge.

Our second example of a downstream task for model embeddings is classifying processes based on formation rules. In this case, we train a classifier using the embedding vectors as features and the rules described in Section~\ref{sec:dataset} as labels. Six \textit{k}-NN classifiers were trained using 5-fold cross-validation and cosine distance. 

\begin{figure}[htpb]
\centering
\includegraphics[width=.85\textwidth]{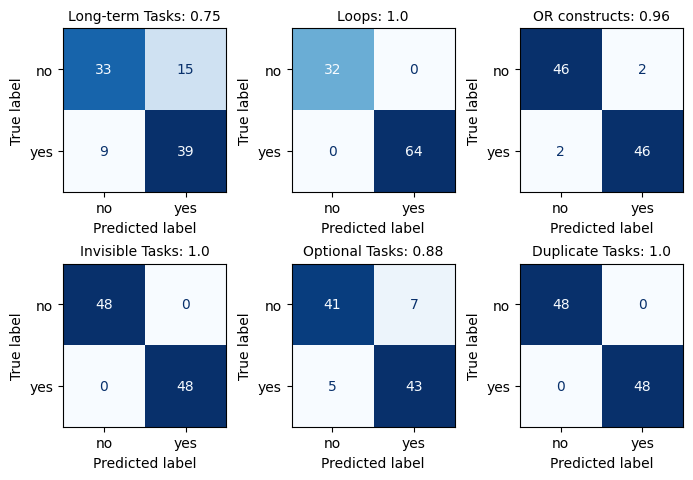}
\caption{Confusion matrices of 1-NN with cosine distance trained to recognize the rules A, B, C, D, E, and F from the model embeddings. Each subfigure title indicates the accuracy achieved using the 5-fold cross-validation methodology.} \label{fig:knn}
\end{figure}

The results can be found in the confusion matrices shown in Figure~\ref{fig:knn}. In this figure, we can observe that some properties of the models, mainly those that achieved higher accuracy, were effectively incorporated by our methodology when creating the embeddings, such as loops, OR-constructs, invisible tasks, and duplicated tasks. However, bypass connections (representing long-term tasks) and optional tasks were more challenging to recognize from the learned embedding, although the results can still be considered satisfactory. This limitation probably arises from the fact that our methodology only considers pairs of consecutive tasks, neglecting, for example, long-term dependencies.

Finally, these examples demonstrate that, in a real-world scenario, our methodology could be used to identify properties of business process models.

\section{Conclusions}\label{sec:conclusions}

In our study, we introduce a novel approach for mapping process models, represented as Petri networks, into embedding vectors. Drawing inspiration from \textit{doc2vec}, our approach inherits the advantageous characteristics of these methods while offering its own simplicity and effectiveness in the realm of Process Mining. Our hypothesis posits that process paths resemble sentences in text documents, allowing for the application of \textit{doc2vec} methodology. This analogy facilitates the learning of vector embeddings for both process models and individual tasks, leading to the development of the PetriNet2Vec methodology.

The training process aims to maximize the probability of task sequences within models. The unsupervised nature of the training, coupled with negative sampling, ensures that robust vector embeddings are learned, capturing the nuances of different process models. Through a series of experiments, we demonstrated that our PetriNet2Vec method was capable of learning the structure of Petri nets, as well as the main properties used in constructing the process models in our dataset. Furthermore, our results showcase the utility of the learned embeddings in two crucial downstream tasks within process mining: process classification and process retrieval.


Further analysis within clusters revealed common formation rules learned by PetriNet2Vec, as depicted by the Decision Tree, indicating strong coherence among cluster members. For instance, examination of cluster $C_0$ (Figure~\ref{fig:cluster_members}) showcased models with shared structural properties, such as the absence of loops, invisible tasks, or duplicated tasks. Additionally, exploring task similarity via cosine similarity matrices provided insights into task relationships in vector space, despite the challenge of interpreting tasks solely by their IDs. By identifying tasks with the highest and lowest similarities to a given task, we gained valuable insights into parallel branches and potential structural relationships within the process flow. 

One limitation we observed during the clustering experiments was that the length of the context window negatively impacts clustering performance. Addressing this issue could enable the characterization of long-range paths within the Petri nets. Furthermore, the PetriNet2Vec model needs to be retrained when applied to new process spaces that utilize a different set of tasks. This retraining requirement poses a challenge in adapting the model to novel process domains without additional training data and computational resources.

The code used in these experiments can be found in our Github repository for reference\footnote{\url{https://github.com/juancolonna/PetriNet2Vec}}. Additionally, our PetriNet2Vec package is now conveniently accessible as a Python package, installable via the \textbf{pip} tool\footnote{\url{https://pypi.org/project/PetriNet2Vec}}. In our future work, we plan to conduct a case study using real-world data collected from a company, as opposed to relying on simulated datasets. This will allow us to validate our findings in a practical, business environment and enhance the applicability of our research. Additionally, we plan to utilize these embeddings to enhance process discovery. For instance, we will compare the effectiveness of our approach with that of a baseline model, aiming to demonstrate improvements in accuracy and efficiency. Furthermore, we plan to expand our methodology by incorporating additional tasks into the model's context. For example, this extension could involve modifying Equation~\ref{equ:maximize} to include $t_{i-1}, t_{i-2}, t_{i-3}, \dots$, thereby integrating a broader task context into the model. This enhancement aims to capture deeper temporal dependencies, potentially improving the model's predictive accuracy and relevance in complex scenarios.

\subsection*{Acknowledgements}
This paper is funded/financed in total by Component 5 - Capitalization and Business Innovation, integrated in the Resilience Dimension of the Recovery and Resilience Plan within the scope of the Recovery and Resilience Mechanism (MRR) of the European Union (EU), framed in the Next Generation EU, for the period 2021 - 2026, within project FAIST, with reference 66. 

It also supported by National Funds through the Portuguese funding agency, FCT - Fundação para a Ciência e a Tecnologia, within project LA/P/0063/2020 (DOI 10.54499/LA/P/0063/2020) under INESC TEC International Visiting Researcher Programme.

This work was supported by the Coordenação de Aperfeiçoamento de Pessoal de Nível Superior - Brazil (CAPES-PROEX) - Code of Financing 001, and also by the Fundação de Amparo à Pesquisa do Estado do Amazonas - FAPEAM - through the POSGRAD 23-24 project. 

\bibliography{references}

\begin{thebibliography}{29}
\expandafter\ifx\csname natexlab\endcsname\relax\def\natexlab#1{#1}\fi
\providecommand{\url}[1]{\texttt{#1}}
\providecommand{\href}[2]{#2}
\providecommand{\path}[1]{#1}
\providecommand{\DOIprefix}{doi:}
\providecommand{\ArXivprefix}{arXiv:}
\providecommand{\URLprefix}{URL: }
\providecommand{\Pubmedprefix}{pmid:}
\providecommand{\doi}[1]{\href{http://dx.doi.org/#1}{\path{#1}}}
\providecommand{\Pubmed}[1]{\href{pmid:#1}{\path{#1}}}
\providecommand{\bibinfo}[2]{#2}
\ifx\xfnm\relax \def\xfnm[#1]{\unskip,\space#1}\fi
\bibitem[{Bae et~al.(2006)Bae, Caverlee, Liu \& Yan}]{bae_process_2006}
\bibinfo{author}{Bae, J.}, \bibinfo{author}{Caverlee, J.}, \bibinfo{author}{Liu, L.}, \& \bibinfo{author}{Yan, H.} (\bibinfo{year}{2006}).
\newblock \bibinfo{title}{Process mining by measuring process block similarity}.
\newblock In {\it \bibinfo{booktitle}{Business {Process} {Management} {Workshops}: {BPM} 2006 {International} {Workshops}, {BPD}, {BPI}, {ENEI}, {GPWW}, {DPM}, semantics4ws, {Vienna}, {Austria}, {September} 4-7, 2006. {Proceedings} 4}\/} (pp. \bibinfo{pages}{141--152}).
\newblock \bibinfo{publisher}{Springer}.
\bibitem[{Cai et~al.(2018)Cai, Zheng \& Chang}]{Cai2018}
\bibinfo{author}{Cai, H.}, \bibinfo{author}{Zheng, V.~W.}, \& \bibinfo{author}{Chang, K.} (\bibinfo{year}{2018}).
\newblock \bibinfo{title}{A comprehensive survey of graph embedding: Problems, techniques, and applications}.
\newblock {\it \bibinfo{journal}{IEEE Transactions on Knowledge and Data Engineering}\/},  {\it \bibinfo{volume}{30}\/}, \bibinfo{pages}{1616--1637}. \DOIprefix\doi{10.1109/TKDE.2018.2807452}.
\bibitem[{Cárdenas~Maita et~al.(2018)Cárdenas~Maita, Martins, López~Paz, Rafferty, Hung, Peres \& Fantinato}]{ana_rocio_cardenas_maita_systematic_2018}
\bibinfo{author}{Cárdenas~Maita, A.~R.}, \bibinfo{author}{Martins, L.~C.}, \bibinfo{author}{López~Paz, C.~R.}, \bibinfo{author}{Rafferty, L.}, \bibinfo{author}{Hung, P. C.~K.}, \bibinfo{author}{Peres, S.~M.}, \& \bibinfo{author}{Fantinato, M.} (\bibinfo{year}{2018}).
\newblock \bibinfo{title}{A systematic mapping study of process mining}.
\newblock {\it \bibinfo{journal}{Enterprise Information Systems}\/},  {\it \bibinfo{volume}{12}\/}, \bibinfo{pages}{505--549}. \URLprefix \url{https://doi.org/10.1080/17517575.2017.1402371}. \DOIprefix\doi{10.1080/17517575.2017.1402371}.
\newblock \bibinfo{note}{Publisher: Taylor \& Francis \_eprint: https://doi.org/10.1080/17517575.2017.1402371}.
\bibitem[{Dijkman et~al.(2011)Dijkman, Dumas, Dongen, Käärik \& Mendling}]{dijkman_similarity_2011}
\bibinfo{author}{Dijkman, R.}, \bibinfo{author}{Dumas, M.}, \bibinfo{author}{Dongen, B.~v.}, \bibinfo{author}{Käärik, R.}, \& \bibinfo{author}{Mendling, J.} (\bibinfo{year}{2011}).
\newblock \bibinfo{title}{Similarity of business process models: {Metrics} and evaluation}.
\newblock {\it \bibinfo{journal}{Information Systems}\/},  {\it \bibinfo{volume}{36}\/}, \bibinfo{pages}{498--516}. \URLprefix \url{https://www.sciencedirect.com/science/article/pii/S0306437910001006}. \DOIprefix\doi{https://doi.org/10.1016/j.is.2010.09.006}.
\bibitem[{Dijkman et~al.(2009)Dijkman, Dumas \& García-Bañuelos}]{dijkman_graph_2009}
\bibinfo{author}{Dijkman, R.}, \bibinfo{author}{Dumas, M.}, \& \bibinfo{author}{García-Bañuelos, L.} (\bibinfo{year}{2009}).
\newblock \bibinfo{title}{Graph matching algorithms for business process model similarity search}.
\newblock In {\it \bibinfo{booktitle}{Business {Process} {Management}: 7th {International} {Conference}, {BPM} 2009, {Ulm}, {Germany}, {September} 8-10, 2009. {Proceedings} 7}\/} (pp. \bibinfo{pages}{48--63}).
\newblock \bibinfo{publisher}{Springer}.
\bibitem[{Dijkman et~al.(2013)Dijkman, Van~Dongen, Dumas, García-Bañuelos, Kunze, Leopold, Mendling, Uba, Weidlich, Weske \& Yan}]{bubenko_short_2013}
\bibinfo{author}{Dijkman, R.~M.}, \bibinfo{author}{Van~Dongen, B.~F.}, \bibinfo{author}{Dumas, M.}, \bibinfo{author}{García-Bañuelos, L.}, \bibinfo{author}{Kunze, M.}, \bibinfo{author}{Leopold, H.}, \bibinfo{author}{Mendling, J.}, \bibinfo{author}{Uba, R.}, \bibinfo{author}{Weidlich, M.}, \bibinfo{author}{Weske, M.}, \& \bibinfo{author}{Yan, Z.} (\bibinfo{year}{2013}).
\newblock \bibinfo{title}{A {Short} {Survey} on {Process} {Model} {Similarity}}.
\newblock In \bibinfo{editor}{J.~Bubenko}, \bibinfo{editor}{J.~Krogstie}, \bibinfo{editor}{O.~Pastor}, \bibinfo{editor}{B.~Pernici}, \bibinfo{editor}{C.~Rolland}, \& \bibinfo{editor}{A.~Sølvberg} (Eds.), {\it \bibinfo{booktitle}{Seminal {Contributions} to {Information} {Systems} {Engineering}}\/} (pp. \bibinfo{pages}{421--427}).
\newblock \bibinfo{address}{Berlin, Heidelberg}: \bibinfo{publisher}{Springer Berlin Heidelberg}.
\newblock \URLprefix \url{http://link.springer.com/10.1007/978-3-642-36926-1_34}. \DOIprefix\doi{10.1007/978-3-642-36926-1_34}.
\bibitem[{Ganesh et~al.(2023)Ganesh, Ramachandran, Varasree, Lakhanpal, Rohini \& Acharjee}]{ganesh_information_2023}
\bibinfo{author}{Ganesh, C.}, \bibinfo{author}{Ramachandran, K.}, \bibinfo{author}{Varasree, B.}, \bibinfo{author}{Lakhanpal, S.}, \bibinfo{author}{Rohini, B.}, \& \bibinfo{author}{Acharjee, P.~B.} (\bibinfo{year}{2023}).
\newblock \bibinfo{title}{Information {Extraction} {Using} {Data} {Mining} {Techniques} {For} {Big} {Data} {Processing} in {Digital} {Marketing} {Platforms}}.
\newblock In {\it \bibinfo{booktitle}{2023 10th {IEEE} {Uttar} {Pradesh} {Section} {International} {Conference} on {Electrical}, {Electronics} and {Computer} {Engineering} ({UPCON})}\/} (pp. \bibinfo{pages}{1662--1667}).
\newblock \bibinfo{publisher}{IEEE} volume~\bibinfo{volume}{10}.
\bibitem[{Hoffmann \& Bergmann(2022)}]{Hoffmann2022}
\bibinfo{author}{Hoffmann, M.}, \& \bibinfo{author}{Bergmann, R.} (\bibinfo{year}{2022}).
\newblock \bibinfo{title}{Using graph embedding techniques in process-oriented case-based reasoning}.
\newblock {\it \bibinfo{journal}{Algorithms}\/},  {\it \bibinfo{volume}{15}\/}. \DOIprefix\doi{10.3390/a15020027}.
\bibitem[{Jurafsky \& Martin()}]{jurafsky_speech_nodate}
\bibinfo{author}{Jurafsky, D.}, \& \bibinfo{author}{Martin, J.~H.} ().
\newblock \bibinfo{title}{Speech and {Language} {Processing}}.
\newblock \URLprefix \url{https://web.stanford.edu/~jurafsky/slp3/}.
\bibitem[{Kunze et~al.(2011)Kunze, Weidlich \& Weske}]{rinderle-ma_behavioral_2011}
\bibinfo{author}{Kunze, M.}, \bibinfo{author}{Weidlich, M.}, \& \bibinfo{author}{Weske, M.} (\bibinfo{year}{2011}).
\newblock \bibinfo{title}{Behavioral {Similarity} – {A} {Proper} {Metric}}.
\newblock In \bibinfo{editor}{S.~Rinderle-Ma}, \bibinfo{editor}{F.~Toumani}, \& \bibinfo{editor}{K.~Wolf} (Eds.), {\it \bibinfo{booktitle}{Business {Process} {Management}}\/} (pp. \bibinfo{pages}{166--181}).
\newblock \bibinfo{address}{Berlin, Heidelberg}: \bibinfo{publisher}{Springer Berlin Heidelberg} volume \bibinfo{volume}{6896}.
\newblock \URLprefix \url{http://link.springer.com/10.1007/978-3-642-23059-2_15}. \DOIprefix\doi{10.1007/978-3-642-23059-2_15} \bibinfo{note}{series Title: Lecture Notes in Computer Science}.
\bibitem[{Le \& Mikolov(2014)}]{Mikolov2014}
\bibinfo{author}{Le, Q.}, \& \bibinfo{author}{Mikolov, T.} (\bibinfo{year}{2014}).
\newblock \bibinfo{title}{Distributed representations of sentences and documents}.
\newblock In {\it \bibinfo{booktitle}{Proceedings of the 31st International Conference on Machine Learning}\/} (pp. \bibinfo{pages}{1188--1196}).
\newblock \bibinfo{publisher}{PMLR} volume~\bibinfo{volume}{32} of {\it \bibinfo{series}{Proceedings of Machine Learning Research}\/}.
\bibitem[{Li et~al.(2008)Li, Reichert \& Wombacher}]{li_measuring_2008}
\bibinfo{author}{Li, C.}, \bibinfo{author}{Reichert, M.}, \& \bibinfo{author}{Wombacher, A.} (\bibinfo{year}{2008}).
\newblock \bibinfo{title}{On measuring process model similarity based on high-level change operations}.
\newblock In {\it \bibinfo{booktitle}{International {Conference} on {Conceptual} {Modeling}}\/} (pp. \bibinfo{pages}{248--264}).
\newblock \bibinfo{publisher}{Springer}.
\bibitem[{McInnes et~al.(2017)McInnes, Healy, Astels et~al.}]{mcinnes2017hdbscan}
\bibinfo{author}{McInnes, L.}, \bibinfo{author}{Healy, J.}, \bibinfo{author}{Astels, S.} et~al. (\bibinfo{year}{2017}).
\newblock \bibinfo{title}{hdbscan: Hierarchical density based clustering.}
\newblock {\it \bibinfo{journal}{J. Open Source Software}\/},  {\it \bibinfo{volume}{2}\/}, \bibinfo{pages}{205}.
\bibitem[{Mikolov et~al.(2013)Mikolov, Chen, Corrado \& Dean}]{Mikolov2013}
\bibinfo{author}{Mikolov, T.}, \bibinfo{author}{Chen, K.}, \bibinfo{author}{Corrado, G.~S.}, \& \bibinfo{author}{Dean, J.} (\bibinfo{year}{2013}).
\newblock \bibinfo{title}{Efficient estimation of word representations in vector space}.
\newblock \URLprefix \url{http://arxiv.org/abs/1301.3781}.
\bibitem[{Montani et~al.(2015)Montani, Leonardi, Quaglini, Cavallini \& Micieli}]{montani_knowledge-intensive_2015}
\bibinfo{author}{Montani, S.}, \bibinfo{author}{Leonardi, G.}, \bibinfo{author}{Quaglini, S.}, \bibinfo{author}{Cavallini, A.}, \& \bibinfo{author}{Micieli, G.} (\bibinfo{year}{2015}).
\newblock \bibinfo{title}{A knowledge-intensive approach to process similarity calculation}.
\newblock {\it \bibinfo{journal}{Expert Systems with Applications}\/},  {\it \bibinfo{volume}{42}\/}, \bibinfo{pages}{4207--4215}. \URLprefix \url{https://linkinghub.elsevier.com/retrieve/pii/S0957417415000421}. \DOIprefix\doi{10.1016/j.eswa.2015.01.027}.
\newblock \bibinfo{note}{Publisher: Elsevier}.
\bibitem[{Murata(1989)}]{Murata1989}
\bibinfo{author}{Murata, T.} (\bibinfo{year}{1989}).
\newblock \bibinfo{title}{Petri nets: Properties, analysis and applications}.
\newblock {\it \bibinfo{journal}{Proceedings of the IEEE}\/},  {\it \bibinfo{volume}{77}\/}, \bibinfo{pages}{541--580}. \DOIprefix\doi{10.1109/5.24143}.
\bibitem[{Narayanan et~al.(2017)Narayanan, Chandramohan, Venkatesan, Chen, Liu \& Jaiswal}]{Annamalai2017}
\bibinfo{author}{Narayanan, A.}, \bibinfo{author}{Chandramohan, M.}, \bibinfo{author}{Venkatesan, R.}, \bibinfo{author}{Chen, L.}, \bibinfo{author}{Liu, Y.}, \& \bibinfo{author}{Jaiswal, S.} (\bibinfo{year}{2017}).
\newblock \bibinfo{title}{graph2vec: Learning distributed representations of graphs}.
\newblock {\it \bibinfo{journal}{CoRR}\/},  {\it \bibinfo{volume}{abs/1707.05005}\/}. \href{http://arxiv.org/abs/1707.05005}{\tt arXiv:1707.05005}.
\bibitem[{Peeva \& van~der Aalst(2023)}]{peeva_grouping_2023}
\bibinfo{author}{Peeva, V.}, \& \bibinfo{author}{van~der Aalst, W.~M.} (\bibinfo{year}{2023}).
\newblock \bibinfo{title}{Grouping {Local} {Process} {Models}}.
\newblock {\it \bibinfo{journal}{arXiv preprint arXiv:2311.03040}\/}, .
\bibitem[{Pismerov \& Pikalov(2022)}]{Pismerov2023}
\bibinfo{author}{Pismerov, A.}, \& \bibinfo{author}{Pikalov, M.} (\bibinfo{year}{2022}).
\newblock \bibinfo{title}{Applying embedding methods to process mining}.
\newblock In {\it \bibinfo{booktitle}{Proceedings of the 5th International Conference on Algorithms, Computing and Artificial Intelligence}\/} ACAI'22.
\newblock \bibinfo{publisher}{Association for Computing Machinery}.
\newblock \DOIprefix\doi{10.1145/3579654.3579730}.
\bibitem[{Recker et~al.(2009)Recker, Rosemann, Indulska \& Green}]{recker_business_2009}
\bibinfo{author}{Recker, J.}, \bibinfo{author}{Rosemann, M.}, \bibinfo{author}{Indulska, M.}, \& \bibinfo{author}{Green, P.} (\bibinfo{year}{2009}).
\newblock \bibinfo{title}{Business process modeling-a comparative analysis}.
\newblock {\it \bibinfo{journal}{Journal of the association for information systems}\/},  {\it \bibinfo{volume}{10}\/}, \bibinfo{pages}{1}.
\bibitem[{{\v R}eh{\r u}{\v r}ek \& Sojka(2010)}]{gensim}
\bibinfo{author}{{\v R}eh{\r u}{\v r}ek, R.}, \& \bibinfo{author}{Sojka, P.} (\bibinfo{year}{2010}).
\newblock \bibinfo{title}{Software framework for topic modelling with large corpora}.
\newblock In {\it \bibinfo{booktitle}{Proceedings of the LREC 2010 Workshop on New Challenges for NLP Frameworks}\/} (pp. \bibinfo{pages}{45--50}).
\newblock \bibinfo{publisher}{ELRA}.
\bibitem[{Syukriilah et~al.(2015)Syukriilah, Kusumo \& Widowati}]{syukriilah_structural_2015}
\bibinfo{author}{Syukriilah, N.}, \bibinfo{author}{Kusumo, D.~S.}, \& \bibinfo{author}{Widowati, S.} (\bibinfo{year}{2015}).
\newblock \bibinfo{title}{Structural similarity analysis of business process model using selective reduce based on {Petri} {Net}}.
\newblock In {\it \bibinfo{booktitle}{2015 3rd {International} {Conference} on {Information} and {Communication} {Technology} ({ICoICT})}\/} (pp. \bibinfo{pages}{1--5}).
\newblock \bibinfo{publisher}{IEEE}.
\bibitem[{Sánchez-Charles et~al.(2016)Sánchez-Charles, Muntés-Mulero, Carmona \& Solé}]{sanchez-charles_process_2016}
\bibinfo{author}{Sánchez-Charles, D.}, \bibinfo{author}{Muntés-Mulero, V.}, \bibinfo{author}{Carmona, J.}, \& \bibinfo{author}{Solé, M.} (\bibinfo{year}{2016}).
\newblock \bibinfo{title}{Process model comparison based on cophenetic distance}.
\newblock In {\it \bibinfo{booktitle}{Business {Process} {Management} {Forum}: {BPM} {Forum} 2016, {Rio} de {Janeiro}, {Brazil}, {September} 18-22, 2016, {Proceedings} 14}\/} (pp. \bibinfo{pages}{141--158}).
\newblock \bibinfo{publisher}{Springer}.
\bibitem[{Van Der~Aalst et~al.(2006)Van Der~Aalst, De~Medeiros \& Weijters}]{hutchison_process_2006}
\bibinfo{author}{Van Der~Aalst, W. M.~P.}, \bibinfo{author}{De~Medeiros, A. K.~A.}, \& \bibinfo{author}{Weijters, A. J. M.~M.} (\bibinfo{year}{2006}).
\newblock \bibinfo{title}{Process {Equivalence}: {Comparing} {Two} {Process} {Models} {Based} on {Observed} {Behavior}}.
\newblock In \bibinfo{editor}{D.~Hutchison}, \bibinfo{editor}{T.~Kanade}, \bibinfo{editor}{J.~Kittler}, \bibinfo{editor}{J.~M. Kleinberg}, \bibinfo{editor}{F.~Mattern}, \bibinfo{editor}{J.~C. Mitchell}, \bibinfo{editor}{M.~Naor}, \bibinfo{editor}{O.~Nierstrasz}, \bibinfo{editor}{C.~Pandu~Rangan}, \bibinfo{editor}{B.~Steffen}, \bibinfo{editor}{M.~Sudan}, \bibinfo{editor}{D.~Terzopoulos}, \bibinfo{editor}{D.~Tygar}, \bibinfo{editor}{M.~Y. Vardi}, \bibinfo{editor}{G.~Weikum}, \bibinfo{editor}{S.~Dustdar}, \bibinfo{editor}{J.~L. Fiadeiro}, \& \bibinfo{editor}{A.~P. Sheth} (Eds.), {\it \bibinfo{booktitle}{Business {Process} {Management}}\/} (pp. \bibinfo{pages}{129--144}).
\newblock \bibinfo{address}{Berlin, Heidelberg}: \bibinfo{publisher}{Springer Berlin Heidelberg} volume \bibinfo{volume}{4102}.
\newblock \URLprefix \url{http://link.springer.com/10.1007/11841760_10}. \DOIprefix\doi{10.1007/11841760_10} \bibinfo{note}{series Title: Lecture Notes in Computer Science}.
\bibitem[{Van~Dongen et~al.(2013)Van~Dongen, Dijkman \& Mendling}]{van_dongen_measuring_2013}
\bibinfo{author}{Van~Dongen, B.}, \bibinfo{author}{Dijkman, R.}, \& \bibinfo{author}{Mendling, J.} (\bibinfo{year}{2013}).
\newblock \bibinfo{title}{Measuring similarity between business process models}.
\newblock {\it \bibinfo{journal}{Seminal Contributions to Information Systems Engineering: 25 Years of CAiSE}\/},  (pp. \bibinfo{pages}{405--419}).
\newblock \bibinfo{note}{Publisher: Springer}.
\bibitem[{Verbeek(2023)}]{PDC2023}
\bibinfo{author}{Verbeek, E.} (\bibinfo{year}{2023}).
\newblock \bibinfo{title}{Process discovery contest}.
\newblock \URLprefix \url{https://icpmconference.org/2023/process-discovery-contest/}.
\bibitem[{Yan et~al.(2012)Yan, Dijkman \& Grefen}]{yan_fast_2012}
\bibinfo{author}{Yan, Z.}, \bibinfo{author}{Dijkman, R.}, \& \bibinfo{author}{Grefen, P.} (\bibinfo{year}{2012}).
\newblock \bibinfo{title}{Fast business process similarity search}.
\newblock {\it \bibinfo{journal}{Distributed and Parallel Databases}\/},  {\it \bibinfo{volume}{30}\/}, \bibinfo{pages}{105--144}.
\newblock \bibinfo{note}{Publisher: Springer}.
\bibitem[{Zeng et~al.(2009)Zeng, Tung, Wang, Feng \& Zhou}]{zeng_comparing_2009}
\bibinfo{author}{Zeng, Z.}, \bibinfo{author}{Tung, A.}, \bibinfo{author}{Wang, J.}, \bibinfo{author}{Feng, J.}, \& \bibinfo{author}{Zhou, L.} (\bibinfo{year}{2009}).
\newblock \bibinfo{title}{Comparing {Stars}: {On} {Approximating} {Graph} {Edit} {Distance}.}
\newblock {\it \bibinfo{journal}{PVLDB}\/},  {\it \bibinfo{volume}{2}\/}, \bibinfo{pages}{25--36}.
\bibitem[{Zhou et~al.(2019)Zhou, Liu, Zeng, Lin \& Duan}]{zhou_comprehensive_2019}
\bibinfo{author}{Zhou, C.}, \bibinfo{author}{Liu, C.}, \bibinfo{author}{Zeng, Q.}, \bibinfo{author}{Lin, Z.}, \& \bibinfo{author}{Duan, H.} (\bibinfo{year}{2019}).
\newblock \bibinfo{title}{A {Comprehensive} {Process} {Similarity} {Measure} {Based} on {Models} and {Logs}}.
\newblock {\it \bibinfo{journal}{IEEE Access}\/},  {\it \bibinfo{volume}{7}\/}, \bibinfo{pages}{69257--69273}. \URLprefix \url{https://ieeexplore.ieee.org/document/8611068/}. \DOIprefix\doi{10.1109/ACCESS.2018.2885819}.

\end{thebibliography}
\end{document}